\documentclass[10pt,twocolumn]{article} 
\usepackage{simpleConference}
\usepackage{times}
\usepackage{graphicx}
\usepackage{amssymb}
\usepackage{url,hyperref}
\usepackage{booktabs}
\usepackage{float}

\graphicspath{{media/}}     

\begin{document}

\title{A Unified Industrial Large Knowledge
Model Framework in Industry 4.0 and Smart
Manufacturing}

\author{Jay Lee, Hanqi Su\thanks{\textit{Corresponding author}} \\
Center for Industrial Artificial Intelligence, Department of Mechanical Engineering, \\
A. James Clark School of Engineering, University of Maryland, College Park, \\
Maryland, United States of America \\
\texttt{\{leejay, hanqisu\}@umd.edu}  \\
}

\maketitle
\thispagestyle{empty}

\begin{abstract}
The recent emergence of large language models (LLMs) demonstrates the potential for artificial general intelligence, revealing new opportunities in Industry 4.0 and smart manufacturing. However, a notable gap exists in applying these LLMs in industry, primarily due to their training on general knowledge rather than domain-specific knowledge. Such specialized domain knowledge is vital for effectively addressing the complex needs of industrial applications. To bridge this gap, this paper proposes a unified industrial large knowledge model (ILKM) framework, emphasizing its potential to revolutionize future industries. In addition, ILKMs and LLMs are compared from eight perspectives. Finally, the “6S Principle” is proposed as the guideline for ILKM development, and several potential opportunities are highlighted for ILKM deployment in Industry 4.0 and smart manufacturing.
\end{abstract}

\section{Introduction}\label{1}

In the era of Industry 4.0~\cite{lasi2014industry}, a paradigm shift is unfolding in the manufacturing sector, driven by the advent of smart manufacturing practices. This revolution has been fueled by advancements in industrial big data analytics~\cite{yan2017industrial}, industrial artificial intelligence (AI)~\cite{lee2018industrial}, machine learning (ML) and deep learning~\cite{sharp2018survey,wang2018deep}, cyber-physical system~\cite {lee2015cyber}, and industrial internet of things~\cite{sisinni2018industrial}. These technologies aim to enhance efficiency, productivity, and flexibility in manufacturing processes.

Recent advances in large language models (LLMs)~\cite{zhao2023survey,chang2024survey} have showcased extraordinary
capabilities in natural language processing, including understanding, interpreting, and generating human language. However, a gap exists in the application of these LLMs within smart manufacturing, primarily because LLMs are predominantly trained on general knowledge, not domain-specific knowledge, which may not be entirely suitable for the specific and complex needs of industrial applications. Therefore, there is an urgent need for the development of an advanced foundation model leveraging the powers of LLMs and domain-specific knowledge to address complex challenges in Industry 4.0 and smart manufacturing.

Recognizing this gap, an industrial large knowledge model (ILKM) framework is proposed for domain-driven, data-centric industrial systems in Industry 4.0 and smart manufacturing. In addition, the “6S Principle” is proposed as a guideline for the development of ILKMs. The role of ILKMs and their comparison with LLMs are discussed in detail. Through this exploration, this paper aims to provide a comprehensive understanding of the transformative power of ILKMs in the modern manufacturing landscape, highlighting their significance and opportunities as a cornerstone of the ongoing industrial revolution.

\begin{figure*}[htbp!]
\centering\includegraphics[width=\linewidth]{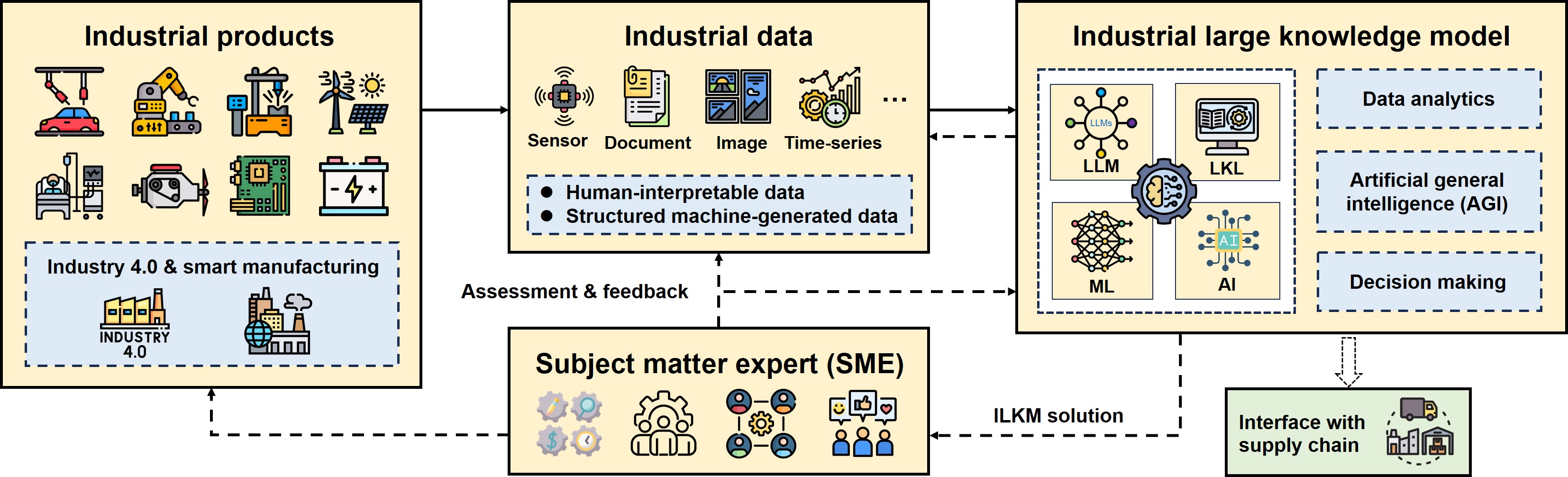}
\caption{General process in Industry 4.0 and smart manufacturing using industrial large knowledge model. Abbreviations: AI: Artificial intelligence; LKL: Large knowledge library; LLM: Large language model; ML: Machine learning.}
\label{fig:1}
\end{figure*}

\section{The role of ILKMs in Industry 4.0 and smart manufacturing}\label{2}

In Industry 4.0 and smart manufacturing, the deployment of ILKMs emerges as a pivotal element. Figure~\ref{fig:1} shows the general process of how ILKM works within Industry 4.0, where ILKM functions at the core of this advanced
manufacturing paradigm. The process begins with the acquisition and management of a vast array of industrial data, derived from diverse industrial products~\cite{raptis2019data,shafiq2019proposition}. This data are categorized into two primary forms: humaninterpretable data and structured machine-generated data. Leveraging technologies such as LLMs, a comprehensive
large knowledge library (LKL), along with various ML and AI techniques~\cite{jan2023artificial}, ILKMs serve as artificial general
intelligence~\cite{goertzel2014artificial}, which plays a vital role in enabling advanced and sophisticated data analytics and problem-solving. These
advanced analytical capabilities, therefore, pave the way for more insightful and informed decision-making processes. Beyond this, ILKMs can also interface with and improve supply chain management~\cite{yandrapalli2023revolutionizing}, leading to more efficient,
resilient, and customer-focused operations. In addition, the solutions generated by ILKMs undergo evaluation by subject matter experts, who play a crucial role in validating and refining the relevant solutions, thereby aiding in the continual optimization of ILKM outputs. This iterative process of assessment and feedback is integral to ensuring the relevance and effectiveness of ILKM solutions. Overall, ILKMs underscore the transformative potential of data-driven approaches, offering detailed and comprehensive
optimization and enhancement directions for industrial products in Industry 4.0 and smart manufacturing.

\begin{figure*}[h!]
\centering\includegraphics[width=\linewidth]{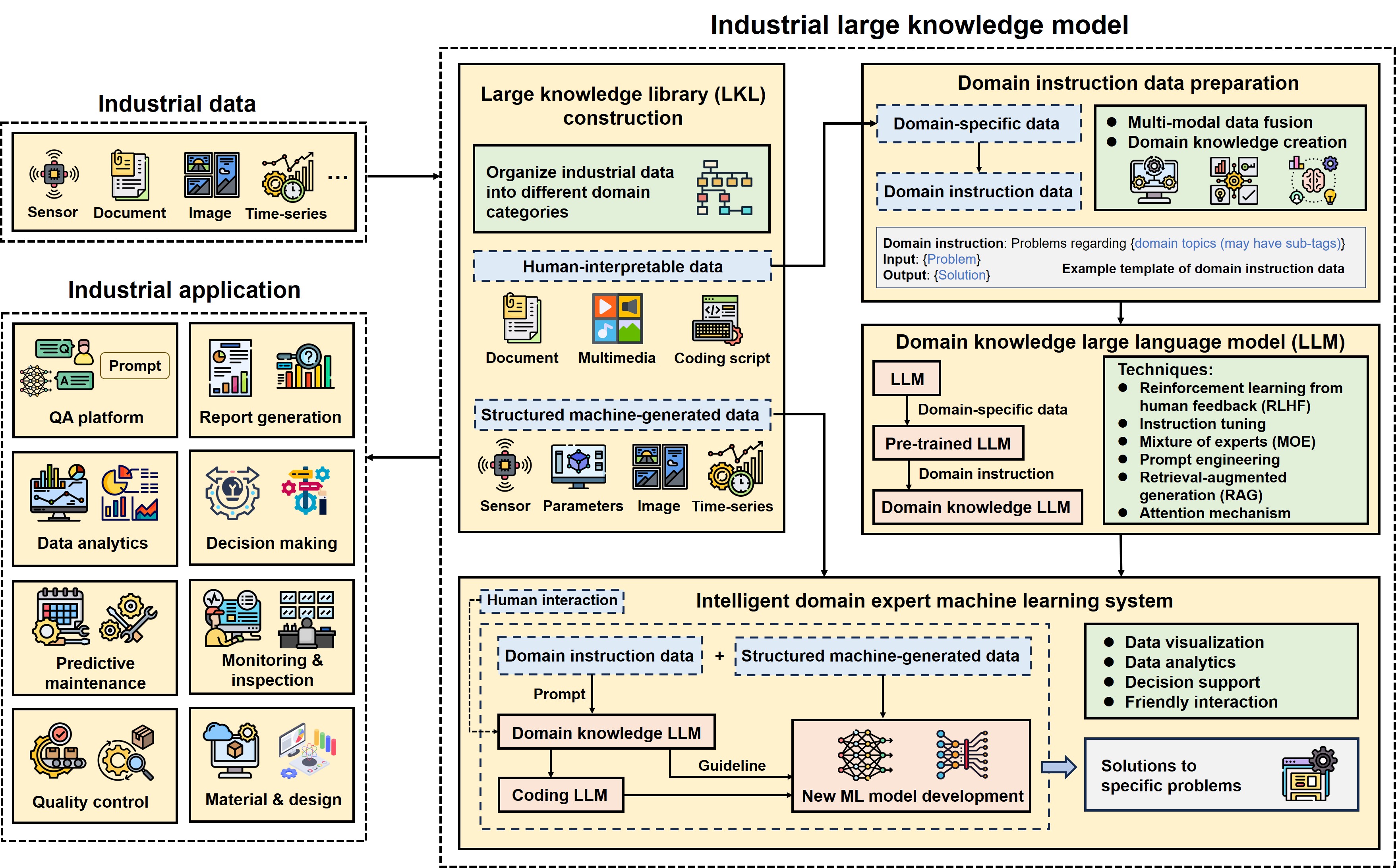}
\caption{A unified industrial large knowledge model framework. Abbreviations: ML: Machine learning; QA: Question answering}
\label{fig:2}
\end{figure*}

\section{ILKM framework}\label{3}
The proposed ILKM framework, shown in Figure~\ref{fig:2}, provides a step-by-step guideline for developing and deploying ILKMs using industrial data to enhance manufacturing capabilities in areas such as predictive maintenance, process optimization, quality control, engineering design, question-answering(QA) platforms, and data analytics. The ILKM framework consists of four pivotal steps: (i) the construction of an LKL categorized by human-interpretable and structured machine-generated data; (ii) the preparation of domain-specific instruction data; (iii) the development of a domain-specific knowledge LLM based on the domain-specific data and domain instruction data; and (iv) the establishment of an intelligent domain expert ML system. As illustrated in Figure~\ref{fig:2}, the details of the ILKM framework are outlined in the subsequent Sections 3.1–3.4.

\subsection{Large knowledge library construction}
The initial step in constructing an ILKM involves the creation of an LKL. This library is pivotal for accommodating the breadth and diversity of industrial data, thus serving as a foundational resource for subsequent analytical tasks. During this phase, it is essential to categorize the data into domain-specific categories systematically. Such an organization enables researchers and data scientists to streamline their efforts, allowing for efficient retrieval of domain-specific data to inform the development of ML models tailored to address distinct industry-related challenges. Within these categories, based on the usage and nature of the data, industrial data can be divided into two primary types: human-interpretable and structured machine-generated. Human-interpretable data, inherently designed for human cognition, comprise elements such as text documents, annotated images, coding scripts crafted by programmers, and multimedia content. This type of data can be seen as insightful information or knowledge and is used for the later development of domain-specific knowledge ML models. On the other hand, structured machine-generated data comprises sensor readings, machine logs, operational parameters, and more. This data type can be leveraged for analytical purposes in technical and industrial contexts.

\subsection{Domain instruction data preparation}
In the second step of the ILKM framework, the focus shifts to transforming domain-specific data (human-interpretable data from LKL) into structured domain instruction sets. This transformation is crucial for enhancing the performance of LLMs in targeted domains by generating domain-centric knowledge and achieving multi-modal data fusion~\cite{lahat2015multimodal}. These structured instructions, vital for finetuning the LLM and retrieving domain knowledge, ensure that the model is proficient in addressing domain-specific
challenges and enhancing problem-solving capabilities~\cite{zhang2023instruction,gao2023retrieval}. As depicted in Figure~\ref{fig:2}, the domain instruction data are organized into three parts: first, the domain instruction, which identifies the problem’s domain and may include sub-tags for refined categorization; second, the input, which clearly outlines the current problem; and third, the output, which presents the corresponding solution.

\subsection{Domain knowledge LLM development}
The third step of the ILKM framework entails an initial pre-training of the base LLM with domain-specific data sourced from LKL. This pre-training imbues the LLM with rich domain-specific knowledge. Following this, the pre-trained LLM undergoes a fine-tuning process, guided by domain instructions, transforming it into a domain knowledge LLM proficient in the designated field. To refine the LLM’s expertise, several common techniques to
enhance and train LLMs can be summarized as follows: reinforcement learning from human feedback~\cite{christiano2017deep}, instruction
tuning~\cite{ouyang2022training,wang2023empower}, mixture of experts~\cite{shazeer2017outrageously}, prompt engineering~\cite{beurer2023prompting,reynolds2021prompt}, retrieval-augmented generation~\cite{gao2023retrieval}, and leveraging attention mechanism~\cite{vaswani2017attention}, The objective of this step is to build a robust LLM that possesses extensive domain knowledge and expertise. This LLM can then guide the development of new ML models capable of addressing complex challenges and real industrial problems.

\begin{table*}[!h]
\caption{Comparison between ILKMs and LLMs. Abbreviations: ILKM: Industrial large knowledge model; LLM: Large language model.} \label{table:1}
\begin{center}
\resizebox{\textwidth}{!}{
\renewcommand{\arraystretch}{1.2}
\begin{tabular}{lll}
\toprule
 &
  ILKM &
  LLM \\
\midrule
Data &
  \begin{tabular}[c]{@{}l@{}}Industrial domain-specific data (human-interpretable data \\ \& structured machine-generated data); private closed source\end{tabular} &
  \begin{tabular}[c]{@{}l@{}}Vast,diverse, and unstructured text data; public open source\end{tabular} \\

Purpose &
  \begin{tabular}[c]{@{}l@{}}Designed for specific industrial tasks; Provide specialized \\ solutions in respective domains\end{tabular} &
  \begin{tabular}[c]{@{}l@{}}Designed for language-related tasks; focus on understanding \\ and generating human language\end{tabular} \\

\begin{tabular}[c]{@{}l@{}}Domain-specific\\ knowledge\end{tabular} &
  \begin{tabular}[c]{@{}l@{}}Specialiized: in-depth, domain knowledge relevant to specific \\ industries\end{tabular} &
  \begin{tabular}[c]{@{}l@{}}General: may lack deep, industry-specific insights\end{tabular} \\

\begin{tabular}[c]{@{}l@{}}Data privacy\\ and security\end{tabular} &
  \begin{tabular}[c]{@{}l@{}}Offer greater control over data privacy and security as they can \\ be hosted within the company's secure environment\end{tabular} &
  \begin{tabular}[c]{@{}l@{}}Potential concerns with data privacy and security as \\ researchers often use licensed pre-trained models developed by \\ other private companies to fine-tune LLMs\end{tabular} \\

\begin{tabular}[c]{@{}l@{}}Integration and\\ customization\end{tabular} &
  \begin{tabular}[c]{@{}l@{}}Tailored and integrated into a growing and evolving industrial\\ ecosystems, aligning with industry-specific needs\end{tabular} &
  \begin{tabular}[c]{@{}l@{}}Need additional resources for integration and customization to \\ fit specific requirements\end{tabular} \\

Scalability &
  \begin{tabular}[c]{@{}l@{}}Adapt and expand based on specific industrial requirements and\\ environment, but need to be balanced with the cost\end{tabular} &
  \begin{tabular}[c]{@{}l@{}}Scalable across platforms but also require significant\\ computational resources\end{tabular} \\

\begin{tabular}[c]{@{}l@{}}Real-time \\ decision making\end{tabular} &
  \begin{tabular}[c]{@{}l@{}}Better suited for real-time decision-making in industrial settings,\\ leveraging specific industry data\end{tabular} &
  \begin{tabular}[c]{@{}l@{}}Limited in handling real-time, complex industrial decision \\ due to generic training\end{tabular} \\

Application &
  \begin{tabular}[c]{@{}l@{}}Process optimization, predictive maintenance, quality control,\\ prognostic and health management, material and design, data \\ analytics, decision-making, question-answering platform, etc\end{tabular} &
  \begin{tabular}[c]{@{}l@{}}Text generation, content creation, conversation, language \\ translation, summarization, etc., Not domain specific\end{tabular} \\
\bottomrule
\end{tabular}}
\end{center}
\end{table*}

\subsection{Intelligent domain expert machine learning system}
Upon the successful training of the domain knowledge LLM, the fourth step involves utilizing it as a domain expert for subsequent specialized model development. In this step, domain instruction data serves as the prompt, propelling the LLM to address specific analytical problems. The domain-specific knowledge LLM, acting on these instruction inputs, proposes targeted solutions. In addition, human experts may interact and intervene, offering strategic guidance to refine the LLM’s outputs. These solutions are then transferred to a coding-focused LLM~\cite{chen2021evaluating,xu2022systematic}, which incrementally develops code aligned with the domain knowledge LLM’s insights, thereby creating a new ML model for specific problems. In addition, the structured machine-generated data serves as the dataset for new ML model training and testing. Finally, this step culminates in the generation of actionable solutions, ready to be integrated into decision-making workflows.

\section{Discussion}\label{4}
This section discusses the comparison between ILKMs and LLMs and introduces the “6S Principle” as a guideline for future ILKM development. It also highlights several potential opportunities for ILKM deployment in Industry 4.0 and smart manufacturing.

\subsection{Comparison between ILKMs and LLMs}
The main difference between ILKMs and LLMs lies in their purpose and functionality. ILKMs are designed to handle specific industrial tasks, utilizing relevant structured industrial data and domain-specific knowledge to provide precise, expert-level solutions. In contrast, LLMs are more generalized, leveraging extensive training on diverse textual data to solve language-related tasks, such as text generation, conversation, and language translation. To better illustrate the characteristics of ILKMs, a detailed
comparison between ILKM and LLM is presented in Table~\ref{table:1}. They are compared and explained from eight perspectives: “Data,” “Purpose,” “Applications,” “Data Privacy and Security,” “Domain-Specific Knowledge,”
“Integration and Customization,” “Scalability,” and “RealTime Decision-Making.”

\begin{figure*}[h!]
\centering\includegraphics[width=\linewidth]{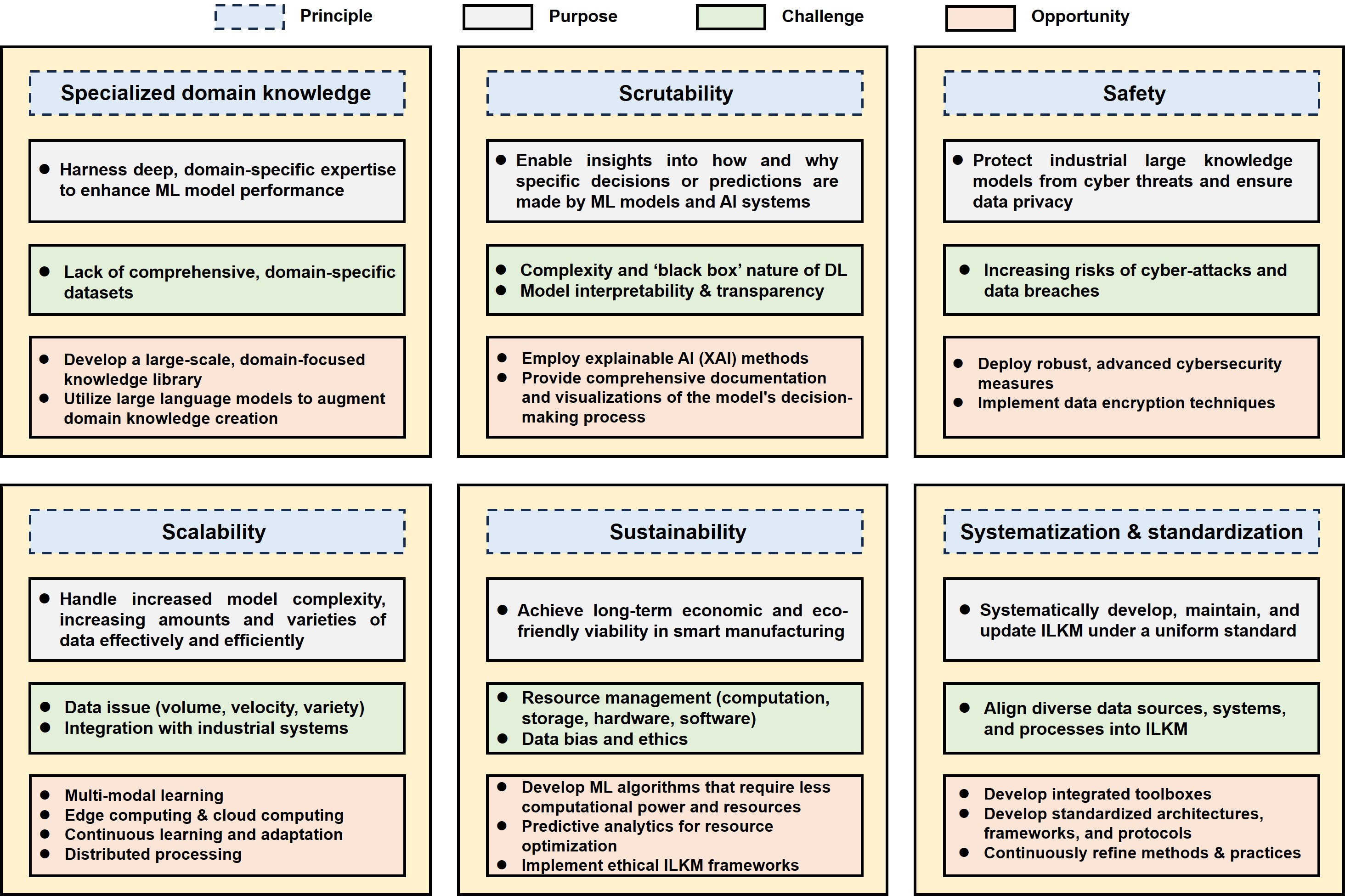}
\caption{The "6S Principl" for ILKM development. 
Abbreviations: AI: Artificial intelligence; DL: Deep learning; ILKM: Industrial large knowledge models; ML: Machine learning}
\label{fig:3}
\end{figure*}

\subsection{Foundational principles for ILKM development}
As illustrated in Figure~\ref{fig:3}, the “6S Principle” is proposed as a guideline for the future development of ILKMs. The “6S Principle” encompasses six key components: “Specialized Domain Knowledge,” “Scrutability,” “Safety,” “Scalability,” “Sustainability,” and “Systematization and Standardization.” The details of the purpose, challenges, and opportunities for each principle are presented in Figure~\ref{fig:3}. All of these principles are crucial for the successful application of ILKMs in industrial settings, ensuring that ILKMs can address specific needs and challenges faced in Industry 4.0 and smart manufacturing.

\subsection{Prospective and opportunity}
There are several opportunities for developing ILKMs in the future of Industry 4.0 and smart manufacturing. In the field of material discovery and synthesis, ILKMs can analyze historical data from the literature to summarize design guidelines and principles as domain-specific instructions. These instructions can then guide the design of new experiments, ultimately facilitating the
discovery of new materials. In the area of engineering design, ILKM can assimilate knowledge from multimodalities (such as text, 2D images, 3D shapes, and sound) gathered from historical products and provide possible optimization directions for designers and engineers to enhance new product performance. In
the realm of prognostics and health management, ILKMs can analyze data from historical failures and maintenance strategies to aid in the diagnosis and
prognosis of complex industrial machine systems, ultimately enabling predictive maintenance and lifecycle management. Furthermore, intelligent QA platforms can be developed across various industrial sectors with the revolution of ILKM. Instead of manual information searches performed by humans, ILKM can automatically retrieve relevant information and generate responses, thus assisting employees with their queries.

\section{Conclusion}\label{5}
This paper presents a unified ILKM framework to address the complex needs of industrial applications by integrating advanced AI, ML, and LLM technologies
with specialized industrial knowledge. The “6S Principle” serves as a foundational guideline for ILKM development, aiming to create a domain-specific, interpretable, secure, scalable, and sustainable ILKM that meets the demands and challenges in Industry 4.0 and smart manufacturing. Moving forward, future research should focus on further integrating cutting-edge AI and ML technologies, continuously refining the framework and its guiding principles based on real-world applications, and leveraging this framework to develop innovative approaches for broader adoption across various industrial sectors. In summary, the ILKM framework shows significant potential for enhancing the intelligence, efficiency, and resilience of future industries.

\bibliographystyle{abbrv}
\bibliography{manufacturing-letters}
\end{document}